\documentclass{article}

\usepackage{arxiv}

\usepackage[utf8]{inputenc} % allow utf-8 input
\usepackage[T1]{fontenc}    % use 8-bit T1 fonts
\usepackage{hyperref}       % hyperlinks
\usepackage{url}            % simple URL typesetting
\usepackage{booktabs}       % professional-quality tables
\usepackage{amsfonts}       % blackboard math symbols
\usepackage{nicefrac}       % compact symbols for 1/2, etc.
\usepackage{microtype}      % microtypography
\usepackage{lipsum}		% Can be removed after putting your text content
\usepackage{graphicx}
\usepackage{natbib}
\usepackage{doi}
\usepackage{algorithm}%
\usepackage{algorithmicx}%
\usepackage{algpseudocode}%
\usepackage{listings}%
\usepackage{amsthm}%
\usepackage{mathrsfs}%
\usepackage{amsmath}

\usepackage{tabularx}

\usepackage{colortbl}
\usepackage[table]{xcolor} % Extended color support for tables
\definecolor{lightorange}{rgb}{1.0, 0.9, 0.4}

\title{Selection of Heart Sound Segments for Synchronous Classification of Multi-channel Heart Sounds}

\date{} 					% Or removing it

\author{
  \textbf{Marcelo Nogueira}\thanks{Corresponding author.} \\
  INESC TEC \& Faculdade de Ci\^encias, Universidade do Porto \\
  \texttt{diogo.m.nogueira@inesctec.pt}
  \And
  \textbf{Jorge H. Oliveira} \\
  LIACC, Universidade da Maia \\
  \texttt{jorgefisicomat@gmail.com}
  \And
  \textbf{Carlos G. Ferreira} \\
  INESC TEC \& Instituto Superior de Engenharia do Porto \\
  \texttt{cgf@isep.ipp.pt}
  \And
  \textbf{Miguel T. Coimbra} \\
  INESC TEC \& Faculdade de Ci\^encias, Universidade do Porto \\
  \texttt{mcoimbra@fc.up.pt}
  \And
  \textbf{Al\'ipio M. Jorge} \\
  INESC TEC \& Faculdade de Ci\^encias, Universidade do Porto \\
  \texttt{amjorge@fc.up.pt}
}

\renewcommand{\shorttitle}{Segment Selection for Synchronous Multi-channel Heart Sound Classification}

\hypersetup{
pdftitle={A template for the arxiv style},
pdfsubject={q-bio.NC, q-bio.QM},
pdfauthor={David S.~Hippocampus, Elias D.~Striatum},
pdfkeywords={First keyword, Second keyword, More},
}

\begin{document}
\maketitle

\begin{abstract}
Cardiac auscultation remains the most cost-effective screening procedure for cardiovascular diseases, and requires listening at the four main auscultation spots. Despite this, automatic heart sound analysis algorithms mostly classify patients using a single heart sound (single-channel), or, when using more than one (multi-channel), analyze each channel individually. To our knowledge, no prior work classifies patients through the synchronous analysis of multi-channel heart sounds, following the procedure used by physicians.
This motivates us to study whether synchronous multi-channel analysis outperforms single-channel approaches, and whether it holds an advantage over asynchronous multi-channel methods that analyze channels one by one, potentially by capturing inter-channel interference phenomena. To answer these questions, we introduce a selection algorithm that identifies optimal heart sound segments from each of the four auscultation spots, which are then fed into a multi-input CNN that classifies patients by analyzing the four selected sounds simultaneously.
Our synchronous approach, combining the proposed selection algorithm with a multi-input CNN, achieves a superior overall accuracy of 96.5\%, a 9.1\% gain over the best-performing single-channel and asynchronous multi-channel methods. The benefit of the proposed segment selection strategy over random selection is confirmed by a paired statistical significance test ($p = 0.003$). These results were obtained on 735 patients from the CirCor DigiScope dataset with complete recordings from all four spots, and their scope and generalizability are discussed in light of this and other methodological considerations.
\end{abstract}


\section{Introduction}\label{sec1}

Cardiovascular Diseases (CVD) can be defined as a heterogeneous set of heart and vessels disorders. According to the World Health Organization, CVD is  the major cause of death worldwide, accounting for 32\% of all deaths globally~\cite{who_2025}.

Many pathological conditions of the cardiovascular system are reflected in some heart-related signals. Among those signals, auscultation of heart sounds is the most cost-effective screening method and also a non-invasive approach widely used in primary CVD diagnosis. However, the accuracy of auscultation is highly dependent on the skills and physician's experience \cite{Barrett_2004}. 
For these reasons, the development of a method that enables reliable low-cost cardiac screening for the general population is highly valuable. 

Standard approaches for developing automatic heart sound analysis algorithms typically comprise four main stages: (i) pre-processing, (ii) segmentation, (iii) feature extraction, and (iv) classification. The pre-processing stage involves applying filtering techniques to mitigate noise and enhance the quality of the heart sounds. Subsequently, segmentation aims to identify the positions and boundaries of the fundamental heart sounds (S1 and S2), thereby delineating the systolic and diastolic periods. 

Feature extraction, the third stage, transforms the raw segmented signal into a compact set of informative and non-redundant features to boost model performance and computational efficiency. While some features are derived from the time domain, the most common and informative ones originate from the frequency and time-frequency domains. Prominent methods for this purpose include the Fourier Transform~\cite{Ranipa_2021}, Mel-Frequency Cepstral Coefficients (MFCCs)~\cite{Bahreini_2025}, and the Discrete Wavelet Transform~\cite{Ahmad_2021}.

Finally, in the classification stage, a classifier is trained on the extracted features to predict the category of each heart sound signal. A variety of classifiers have been employed for this task based on such feature sets~\cite{Ranipa_2021, Nogueira_BHI, Bahreini_2025}.

Recently, deep learning has become the state-of-the-art for abnormal heart sound detection. Convolutional Neural Networks (CNNs) have proven particularly successful, applied either directly to segmented time-domain signals or to their time-frequency representations \cite{Nogueira_BHI}. For instance, \citeauthor{Ranipa_2021} \cite{Ranipa_2021} proposed a multimodal CNN fusion architecture for phonocardiogram (PCG) signal classification. Their approach extracts three distinct feature types: MFCCs, Mel-spectrograms, and frequency-domain features (chroma and spectral contrast). This method achieved a top accuracy of 98\% in abnormality detection across several experimental implementations. However, the generalizability of these results may be limited. The dataset used in their study \cite{Son_2018} was curated by combining signals from approximately 50 online sources and excluding noisy recordings. This selective process may result in a dataset that is not fully representative of real-world acoustic conditions.

\citeauthor{Bahreini_2025} \cite{Bahreini_2025} introduced a hybrid method for heart sound classification that fuses deep learning and handcrafted features. They extracted deep features from time-frequency images using a pre-trained VGG-16 network and handcrafted MFCC features from Empirical Wavelet Transform (EWT) sub-bands. These feature sets were combined using Canonical Correlation Analysis (CCA) to create a compact, discriminative feature vector. Their model, evaluated on the PhysioNet 2016 dataset, achieved 99.55\% accuracy in classifying five cardiac conditions and demonstrated notable robustness to noise.

\citeauthor{Fang_2024} \cite{Fang_2024} introduced a novel method for heart sound classification that combines a Frequency-Balanced Power Spectral Intensity (FBPSI) envelope with a Multi-Level Feature Encoding (MLFE) algorithm. The FBPSI is designed to minimize differences between frequency bands, providing a more effective spectral representation. The MLFE algorithm then decomposes this envelope using the Maximum Overlap Discrete Wavelet Transform (MODWT) and encodes it with overlapping windows to generate a discriminative 128-dimensional feature vector. These features are classified using an Ensemble Bagged-Tree Classifier. Evaluated on the PhysioNet 2016 dataset, their method achieved a high accuracy of 98.73\% for binary classification (normal vs. abnormal) and 98.12\% for a ternary task (normal vs. two types of Hypertrophic Cardiomyopathy) on a self-collected dataset, demonstrating robust performance across different clinical scenarios.

%\citeauthor{Ahmad_2021} \cite{Ahmad_2021} proposed a heart sound classification algorithm based on a Long Short-Term Memory (LSTM) network. The model utilized features from both the time and frequency domains, specifically the Discrete Wavelet Transform (DWT) and MFCCs. The authors conducted five experiments to evaluate different feature groups and feature selection techniques. The best performance, achieving a final score of 90\%, was obtained using a combination of random over-sampling and feature selection. All experiments were evaluated on the PhysioNet/CinC Challenge 2016 database \cite{Liu_2016}.

\citeauthor{Nogueira_BHI} \cite{Nogueira_BHI} investigated the impact of multi-channel heart sound analysis on patient classification using the CirCor Digiscope Dataset \cite{CirCor_Dataset}. Their methodology involved segmenting the original PCG signals into three-second intervals and extracting MFCCs as features. Each segment was first classified as normal or abnormal, and these segment-level predictions were then aggregated to produce a final classification for the entire PCG signal and, consequently, the patient. The authors compared two approaches: a single-channel method, which classifies the patient based on one heart sound, and a multi-channel method, which combines predictions from sounds collected at the four standard auscultation spots. The XGBoost algorithm yielded the best single-channel performance at 82.5\% overall accuracy. In contrast, the multi-channel approach, which fused predictions from all four channels, achieved a superior overall accuracy of 87.4\% using a Support Vector Machine (SVM). This result demonstrates a clear performance gain from leveraging multi-channel information.

Recently, the multi-input CNNs architecture has gained popularity for classification tasks across diverse fields such as image processing, remote sensing, medical diagnosis, and agriculture \cite{MAZUMDER_2022, SANCHEZCAUCE_2021}. Unlike a standard CNN, which processes a single input channel, a multi-input CNN can simultaneously process multiple streams of data. This architecture employs parallel input branches, each dedicated to a distinct data modality (e.g., text, images, or numerical data). Each branch is processed by independent convolutional layers, and the resulting feature representations are merged at a later stage to form a unified output. By learning complex relationships between different modalities, this approach often achieves superior performance compared to single-input models.

Cardiac auscultation, as performed by healthcare professionals, relies on listening to heart sounds from four primary locations: the aortic (AV), pulmonary (PV), tricuspid (TV), and mitral (MV) valves \cite{McGee_2018}. A significant obstacle in developing automated heart sound analysis algorithms has been the scarcity of large, publicly available datasets. Until recently, only two major public datasets existed: The Pascal Challenge Database~\cite{Gomes_2013} and The PhysioNet/CinC Challenge 2016 Database~\cite{Liu_2016}. Crucially, these datasets lack synchronous recordings from all four auscultation spots, making it impossible to develop algorithms that mimic the comprehensive, multi-channel approach of a clinical examination.

The recent release of the CirCor Digiscope Dataset~\cite{CirCor_Dataset} presents a unique opportunity. It provides, for most patients, one or more heart sound recordings per auscultation spot, all screened for the presence of murmurs. Motivated by this rich data source, we leverage the multi-input CNN architecture to develop a novel patient classification model. This model performs a synchronous analysis of data from all four auscultation spots, mirroring the physician's holistic approach.

This work aims to address three key questions. First, we investigate whether synchronous multi-channel analysis yields superior pathology detection performance compared to single-channel approaches. Second, we examine if synchronous analysis holds an advantage over asynchronous multi-channel methods, potentially by capturing inter-channel interference phenomena. Finally, given that our model relies on three-second signal segments for each auscultation spot, we plan to investigate whether applying a representative segment selection algorithm can enhance the overall accuracy and robustness of the classification models.

The rest of the paper is organized as follows. Section II details the dataset, while Section III describes the proposed methodology. Section IV presents the experimental setup and the evaluation metrics, and Section V discusses the experimental results. Finally, Section VI provides the concluding remarks.

\section{The PCG Dataset}

We developed and validated our work using the CirCor DigiScope dataset \cite{CirCor_Dataset}. This dataset comprises 5,277 heart sound recordings collected asynchronously from the four main auscultation spots (AV, PV, MV, TV) during mass screening campaigns in Brazil involving 1,568 participants. The cohort includes 75\% of individuals with normal heartbeats and 25\% with diagnosed cardiac conditions. Recorded in an ambulatory environment, the dataset contains various noisy components, making it a representative sample of real-world conditions for developing robust automatic heart sound analysis algorithms. Further details are available in the original publication \cite{CirCor_Dataset}.

From the total of 1,568 patients in the dataset, 735 met the inclusion criterion of having a complete set of recordings from
all four auscultation spots and were eligible for our multi-input CNN model. The remaining 833 patients were excluded due to
incomplete data (i.e., recordings from only one, two, or three spots), which does not fulfill the model's input requirements.
This inclusion criterion follows directly from our objective of mirroring standard clinical practice, in which a complete cardiac auscultation examination requires listening at all four primary locations~\cite{McGee_2018}; a patient classification model intended to reflect this procedure must therefore be trained and evaluated on patients for whom all four corresponding recordings are available. We acknowledge that this requirement reduces the number of patients available for training and evaluation relative to single-channel or asynchronous multi-channel approaches, which can make use of partial recordings; this trade-off between fidelity to clinical procedure and sample size is revisited in the Discussion.

\section{Methodology}

\begin{figure}
\centering
\includegraphics[width=1\linewidth]{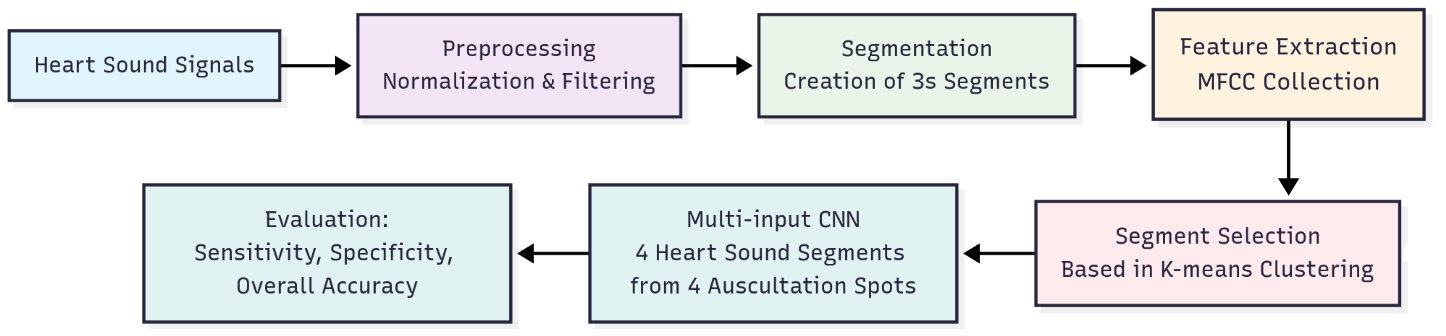}
\caption{\label{fig:Diagram_methodology} Diagram that summarizes the methodology followed to implement the proposed model.}
\end{figure}

This work investigates whether the synchronous analysis of heart sounds from the four auscultation spots can capture inter-channel correlations to improve patient outcome prediction. To test this hypothesis, we developed a multi-channel model based on the methodology summarized in Figure~\ref{fig:Diagram_methodology}:

\begin{itemize}
    \item \textbf{Pre-processing:} Noise removal and signal enhancement (Section~\ref{Pre-Processing_section}).
    \item \textbf{Segmentation:} Division of signals into three-second segments using provided annotations for data augmentation (Section~\ref{Segmentation_section}).
    \item \textbf{Feature Extraction:} Computation of Mel-Frequency Cepstral Coefficients (MFCCs) from the segmented signals (Section~\ref{Feature_Extration_Section}).
    \item \textbf{Segment Selection:} Identification of the most representative three-second segment per auscultation spot for model input (Section~\ref{Segment_selection_Section}).
    \item \textbf{Model Training:} Training of multi-channel and single-channel classifiers (Section~\ref{Classifier}).
    \item \textbf{Model Evaluation:} Performance assessment of the proposed models (Section~\ref{Results_section}).
\end{itemize}

\subsection{Pre-processing}
\label{Pre-Processing_section}

Heart sound signals were sampled at 4KHz and with a 16-bits resolution. The principal frequency components of heart sounds and murmurs are between 20 and 700Hz, and higher frequencies do not have major clinical significance ~\cite{McGee_2018}. Hence, a band-pass Butterworth filter with cutoff frequencies at 20 - 750 Hz was used~\cite{McGee_2018}. The amplitude of the filtered signal was then normalized between -1 and +1.

\subsection{Segmentation}
\label{Segmentation_section}

\begin{figure}
\centering
\includegraphics[width=0.9\linewidth]{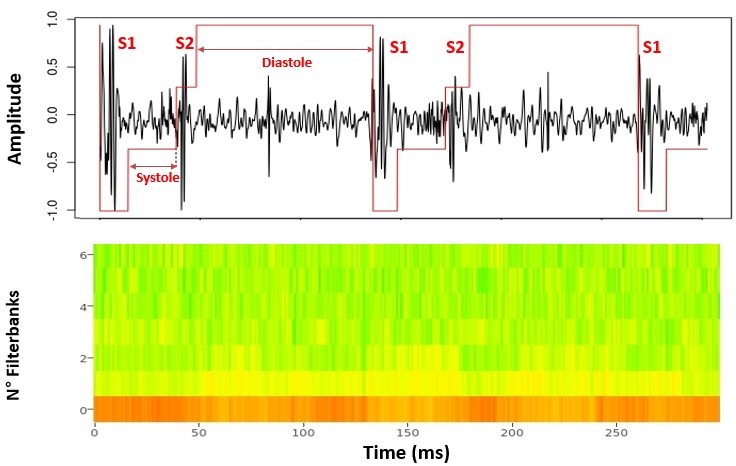}
\caption{\label{fig:PCG_MFCC_image} At the top of the figure we have an example of a abnormal PCG signal with the fundamental heart sounds states delimited (red line). At the bottom of the figure, an example of the MFCC features collected.}
\end{figure}

Leveraging the fundamental heart sound (S1, S2) boundary annotations provided in the CirCor Digiscope Dataset~\cite{CirCor_Dataset}, the original PCG signals were segmented. Segments of length T = 3 seconds were extracted starting from the S1 state, ensuring each segment contained at least one full cardiac cycle. This segment length is consistent with recent state-of-the-art PCG classification methods~\cite{Nogueira_BHI}. From a total of 2,940 PCG signals, this process generated 49,833 three-second segments. An example of the resulting segments is illustrated in Figure~\ref{fig:PCG_MFCC_image}.

\subsection{Feature Extraction}
\label{Feature_Extration_Section}

Following segmentation, MFCCs were extracted from the three-second segments as the primary feature set. MFCCs were selected for their well-established effectiveness in audio processing, particularly due to their alignment with human auditory perception \cite{Nogueira_2017}. This makes them a suitable choice for replicating the diagnostic process of human experts, and their success in heart sound classification has been demonstrated in prior work \cite{Ranipa_2021, Bahreini_2025, Nogueira_BHI}.

The extraction process applied overlapping sliding windows to each segment, using a 25 ms window length with a 10 ms step size. A total of 6 MFCCs were computed per window. Given the 300 time frames in a three-second segment, this procedure yielded a feature matrix of size 6 x 300 for each segment. An example of the resulting MFCCs is shown at the bottom of Figure~\ref{fig:PCG_MFCC_image}.

\subsection{Segment Selection}
\label{Segment_selection_Section}

\begin{figure}
\centering
\includegraphics[width=0.8\linewidth]{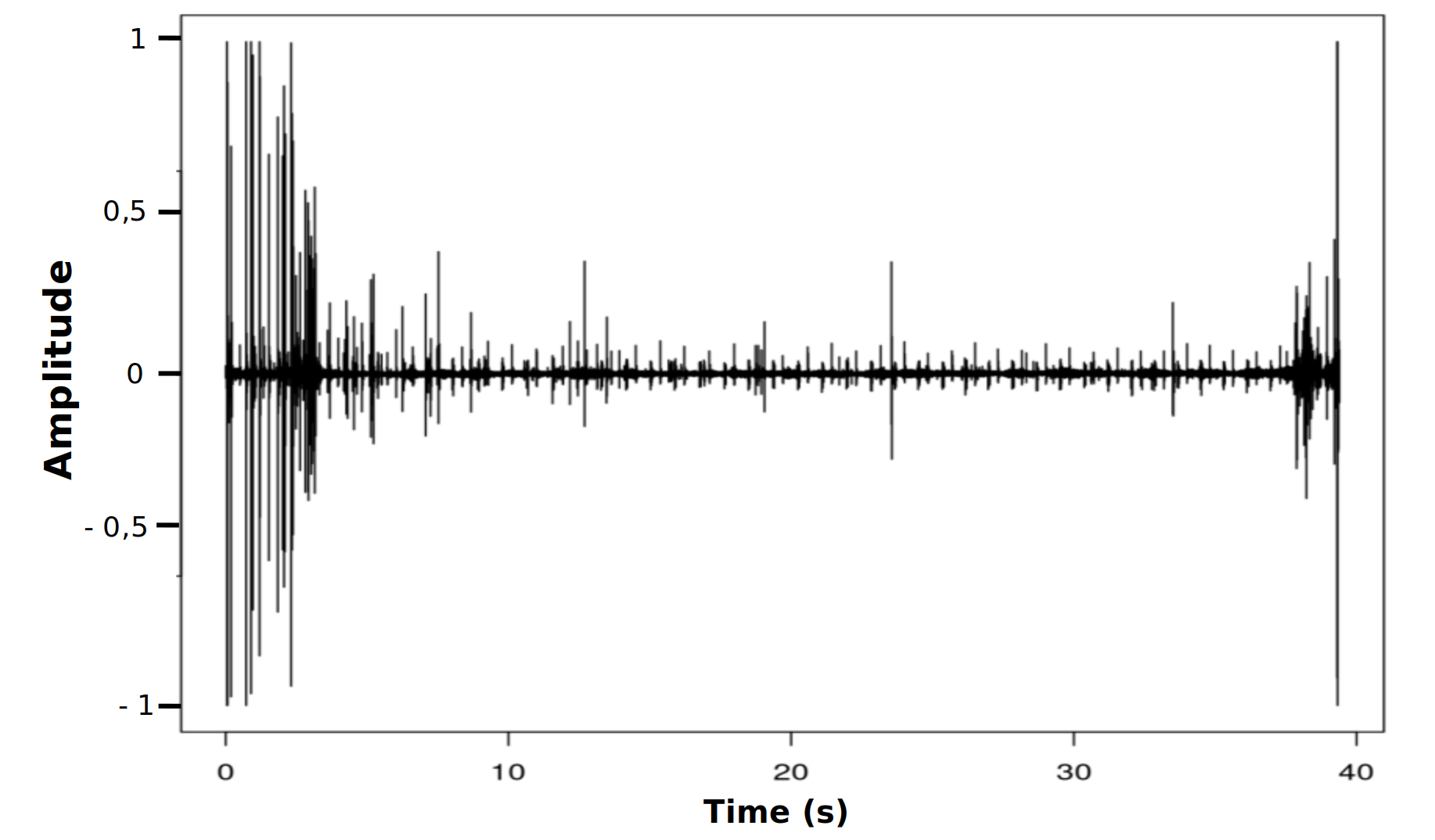}
\caption{\label{fig:noisy_sound} Example of an original heart sound segment, before the segmentation in segments of three seconds.}
\end{figure}

\begin{figure}
\centering
\includegraphics[width=0.8\linewidth]{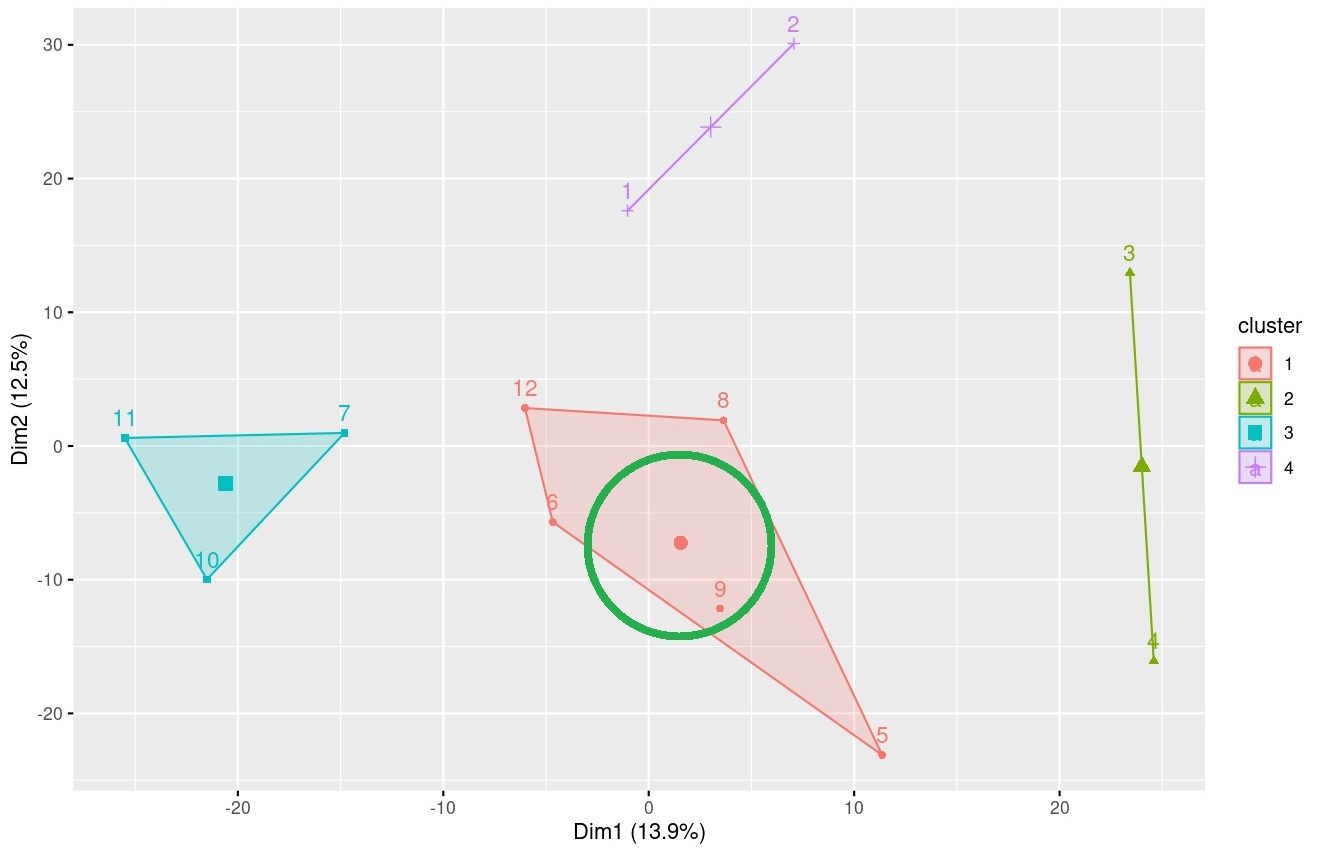}
\caption{\label{fig:Cluster_image} Example of the clustering algorithm used to select the most representative segment of each heart sound.}
\end{figure}

The proposed multi-channel model requires one three-second segment from each of the four auscultation spots as input. To maximize patient inclusion—especially for cases with limited segments per spot—we employ a single segment per spot. However, as most heart sounds yield multiple segments after segmentation, a robust selection method is needed to avoid noisy or unrepresentative portions. Noise, often caused by stethoscope movement (see Figure~\ref{fig:noisy_sound}), frequently contaminates the beginning, end, or middle of recordings, even after pre-processing.

To address this, we applied a K-means clustering algorithm~\cite{MacQueen_1967} to select the most representative segment for each heart sound. The process is as follows:
\begin{enumerate}
    \item Determine the optimal number of clusters \(K\) using Silhouette Analysis.
    \item Group all segments from one heart sound into \(K\) clusters.
    \item Select the largest cluster.
    \item Compute the centroid of the selected cluster.
    \item Choose the segment closest to the centroid as the representative sample.
\end{enumerate}

This method is illustrated in Figure~\ref{fig:Cluster_image}, where 12 segments from one PCG signal are grouped into 4 clusters. Cluster 1, the largest, is chosen, and segment 9 is selected for its proximity to the centroid. By selecting the largest cluster, we ensure the chosen segment is representative of the patient's typical heart sound, minimizing the influence of outliers or noisy segments.

Thus, each patient is represented by four selected segments—one per auscultation spot—forming the input to our multi-input CNN. This approach not only mitigates noise but also simplifies the model compared to prior work~\cite{Nogueira_2017, Gomes_2013}, which often requires dozens of segments per patient for classification.

The most representative segment selection method we just described is summarized in Algorithm \ref{alg:segment_selection}.

\begin{algorithm}
    \caption{\label{alg:segment_selection}Algorithm to select the representative segment}
    \begin{algorithmic}[1]  
        \Require Segments: A list of 3-second heart sound segments
        \Ensure Selection of the most representative segment
        
        \vspace{0.9em}
        \State \textbf{Step 1:} Determine the ideal number of clusters $K$ using silhouette analysis
        \State $K \gets \text{silhouette\_analysis(segments)}$
        
        \vspace{0.9em}
        \State \textbf{Step 2:} Organize segments into $K$ clusters
        \State $\text{clusters} \gets \text{cluster\_segments(segments, K)}$

        \vspace{0.9em}
        \State \textbf{Step 3:} Identify the largest cluster
        \State $\text{largest\_cluster} \gets \text{find\_largest\_cluster(clusters)}$

        \vspace{0.9em}
        \State \textbf{Step 4:} Calculate the centroid of the largest cluster
        \State $\text{centroid} \gets \text{calculate\_centroid(largest\_cluster)}$

        \vspace{0.9em}
        \State \textbf{Step 5:} Compute distances from each sample to the centroid
        \State $\text{distances} \gets []$
        \For{each sample in largest\_cluster}
            \State $\text{distance} \gets \text{calculate\_distance(sample, centroid)}$
            \State $\text{distances.append((sample, distance))}$
        \EndFor

        \vspace{0.9em}
        \State \textbf{Step 6:} Select the sample with the minimum distance to the centroid
        \State $\text{representative\_segment} \gets \text{sample\_with\_min\_distance(distances)}$

        \vspace{0.9em}
        \State \Return $\text{representative\_segment}$
    \end{algorithmic}
\end{algorithm}

The clustering-based selection strategy was compared against random segment selection, the most common approach in the absence
of an explicit selection criterion. Simpler heuristics, such as selecting the segment with the highest estimated signal-to-noise ratio or the temporally central segment of each recording, were not evaluated in this work, but constitute reasonable alternatives that could further clarify the specific contribution of cluster-based cohesion, as opposed to any principled selection criterion, to the reported performance gain.

\subsection{CNN multi-input Classifier}
\label{Classifier}

\begin{figure*}
\centering
\includegraphics[width=0.55\textwidth]{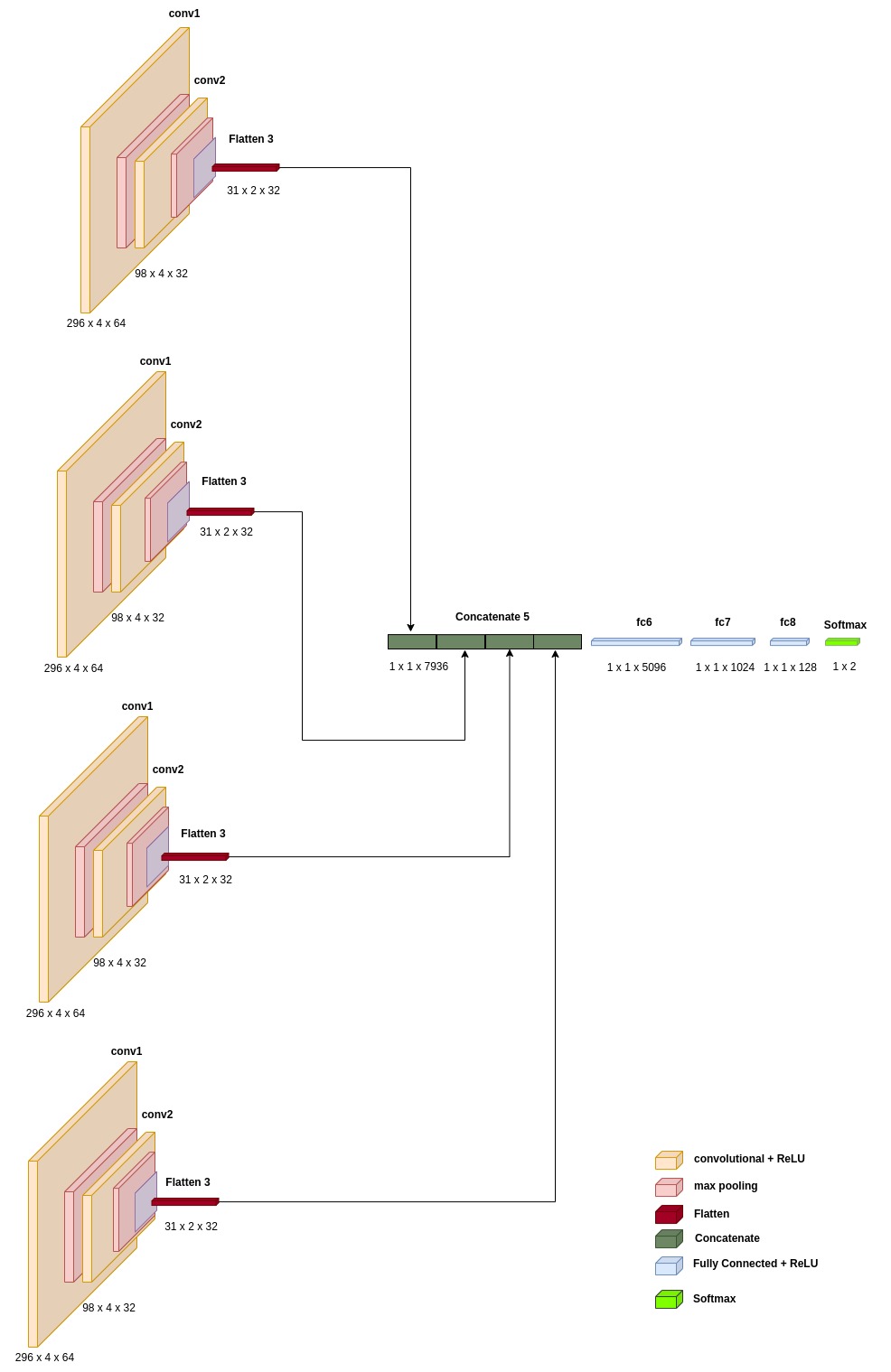}
\caption{\label{fig:Arquitetura_CNN} Overview of our multi-input CNN model.}
\end{figure*}

In this section, we introduce the overall architecture of the proposed model. Figure \ref{fig:Arquitetura_CNN} shows the architecture of our multi-input CNN model for patients classification.

This section details the architecture of the proposed multi-input CNN model for patient classification, illustrated in Figure~\ref{fig:Arquitetura_CNN}. The model takes as input the MFCC matrices of four three-second segments—one from each auscultation spot—each fed into a separate parallel branch.

Each branch processes its input through two convolutional layers (with 64 and 32 filters of size \(5 \times 1\) and \(5 \times 2\), respectively, using ReLU activation) followed by \(3 \times 1\) max-pooling layers. The resulting feature maps are then flattened into 1D vectors. The four vectors are concatenated and passed through a series of fully connected layers (sizes 5096, 1024, 128, and 2), with the final layer employing a softmax activation for classification.

The network architecture was determined empirically. Various configurations were tested, ranging from millions to tens of millions of parameters, without significant performance differences. Consequently, the simplest effective architecture was selected, totaling 9,287,426 parameters.

The model was trained using ten-fold stratified cross-validation. Training data was shuffled and processed in batches of size 10, with 10\% held out for validation. We used the Adam optimizer with a learning rate of 0.0001 over 150 epochs. The model with the lowest categorical cross-entropy loss on the validation set was retained for final testing.

\section{Experimental Setup and Evaluation Metrics}

This section describes the experimental protocol used to assess the proposed method and baseline approaches. The study utilized a dataset of 735 patients, each represented by four three-second signal segments—one from each auscultation spot. Each segment was converted into a 300×6 MFCC matrix, serving as input for the classification network.

Given the class imbalance ($\approx$ 80\% normal vs. 20\% abnormal patients), we employed patient-level ten-fold stratified cross-validation to maintain the original class distribution across all folds.

To evaluate the contributions outlined in Section~\ref{sec1}, we compared our proposed model against three baseline approaches using the same 735 patients:

\begin{itemize}
    \item \textbf{Single-channel approach}: This common method uses only one heart sound per patient for classification. For each patient, we selected the longest available sound to maximize the number of three-second segments after segmentation, resulting in 16,756 segments from 735 original sounds. We applied SVM, Random Forest, XGBoost, and CNN classifiers at the segment level. Patient classification was determined by a voting rule: if any segment from a patient's sound was classified as abnormal, the patient was labeled abnormal.

    \item \textbf{Asynchronous multi-channel approach}: This method uses all four heart sounds per patient but processes them independently. The 2,940 sounds (4 per patient) were segmented into 49,833 three-second segments. The same classifiers were applied at the segment level, and patient diagnosis was determined by a majority vote across the four sounds: if any sound was classified as abnormal, the patient was considered abnormal.

    \item \textbf{Multi-input CNN without segment selection}: This variant of our model performs synchronous analysis of four heart sounds but uses randomly selected segments instead of the proposed representative segment selection method.
\end{itemize}

\subsection{Evaluation Metrics}

Model performance was evaluated using sensitivity (Se), specificity (Sp), and overall accuracy, defined in Equations~\ref{eq:sensitivity}–\ref{eq:overall} based on true positives (TP), true negatives (TN), false positives (FP), and false negatives (FN):

\begin{equation}
Se = \frac{TP}{TP+FN}
\label{eq:sensitivity}
\end{equation}

\begin{equation}
Sp = \frac{TN}{TN+FP}
\label{eq:specificity}
\end{equation}

\begin{equation}
Overall = \frac{Se + Sp}{2}
\label{eq:overall}
\end{equation}

These metrics were selected for comparative purposes, as they align with the evaluation standards used in most studies leveraging the PhysioNet/CinC Challenge 2016 Database \cite{Liu_2016}.

\section{Results and Discussion}
\label{Results_section}

Overall, the multi-input CNN model with segment selection achieves state-of-the-art performance with an overall accuracy of 96.54\%, establishing a new benchmark for synchronous multi-channel heart sound classification, as no prior work has explored this approach. Complete results are shown in Table~\ref{results}.

\begin{table*}[h]
\renewcommand{\arraystretch}{1.5}
\small\centering
\caption{Results of the experiments carried out.}\label{results}%
\begin{tabularx}{\textwidth}{@{}X l l l l@{}}
\toprule
{\bfseries Experiment} & {\bfseries Classifier}  & {\bfseries Sensitivity} & {\bfseries Specificity} & {\bfseries Overall}\\
\midrule
Single-channel & SVM & 0.940 & 0.698 & 0.819 (0.062) \\
Single-channel & RF  & 0.987 & 0.426 & 0.707 (0.070) \\
\rowcolor[gray]{0.9}
Single-channel & XGBoost & 0.926 & 0.724 & 0.825 (0.060) \\
Single-channel & CNN & 0.933 & 0.670 & 0.802 (0.058) \\
\midrule
\rowcolor[gray]{0.9}
Asynchronous multi-channel & SVM & 0.947 & 0.802 & 0.875 (0.039) \\
Asynchronous multi-channel & RF  & 0.755 & 0.960 & 0.857 (0.070) \\
Asynchronous multi-channel & XGBoost & 0.960 & 0.781 & 0.870 (0.044) \\
Asynchronous multi-channel & CNN & 0.955 & 0.793 & 0.874 (0.049) \\
\midrule
Multi-input CNN without Segments Selection & CNN & 0.873 & 0.982 & 0.928 (0.066) \\
\rowcolor{orange!20}
Multi-input CNN & CNN & 0.955 & 0.976 & 0.965 (0.043) \\
\bottomrule
\end{tabularx}
\end{table*}

Among single-channel methods, XGBoost achieved the highest accuracy at 82.5\%. The multi-input CNN's superior performance (a 14-point gain) demonstrates a clear advantage of leveraging multi-perspective data from different auscultation spots. This confirms our first hypothesis from Section~\ref{sec1}: synchronous multi-channel analysis significantly improves pathology detection compared to single-channel approaches. We attribute this gain to the model's ability to capture murmur-related interference patterns across channels, which single-channel architectures cannot detect.

For asynchronous multi-channel approaches, SVM yielded the best result at 87.5\%, outperforming all single-channel methods and supporting the general advantage of multi-channel information. However, it underperformed compared to our synchronous multi-input CNN (87.5\% vs. 96.54\%). Since both methods use the same four heart sounds, the performance gap suggests that synchronous analysis enables the capture of inter-channel interference phenomena that asynchronous approaches miss, addressing our second research question.

It should be noted that the multi-input CNN and the single-channel and asynchronous baselines differ not only in how auscultation spots are combined, but also in model capacity: the multi-input architecture jointly processes four parallel convolutional branches, whereas the baseline classifiers operate on a single MFCC representation at a time. Part of the observed performance gain may therefore also reflect the increased representational capacity of the multi-input architecture itself, rather than synchronous fusion alone. Disentangling these two factors, for instance by comparing the multi-input CNN against a single-input CNN of comparable parameter count applied to a naive concatenation of the four channels, would help isolate the specific contribution of synchronous fusion, and is identified as a direction for future work. Nevertheless, the consistent ordering of results across all four experimental configurations tested (single-channel, asynchronous multi-channel, and synchronous multi-channel with and without segment selection) supports the interpretation that combining information from multiple auscultation spots, and doing so synchronously, both contribute positively to classification
performance.

The segment selection process further enhanced the multi-input CNN's performance, providing a 3.7-point boost over random segment selection (92.8\% vs. 96.5\%). This confirms our third hypothesis: careful segment selection reduces inter-segment variability and minimizes noise, positively impacting model accuracy.

To assess whether this gain reflects a genuine effect of the segment selection strategy rather than fold-level variability, a
paired Wilcoxon signed-rank test was conducted across the 10 paired cross-validation folds underlying the results reported for
the multi-input CNN with and without segment selection. Cluster-based selection achieved a significantly higher overall score than random selection ($p = 0.003$; paired $t$-test: $p = 0.004$), with a moderate effect size (Cohen's $d = 0.49$), indicating that the observed improvement is unlikely to be attributable to chance alone.

Notably, even without segment selection, the multi-input CNN (92.8\%) substantially outperformed both single-channel and
asynchronous multi-channel approaches, underscoring the inherent advantage of synchronous multi-channel analysis.

Finally, the multi-input CNN exhibited one of the lowest standard deviations in 10-fold cross-validation (Table~\ref{results}), indicating greater robustness and consistency compared to other methods, particularly single-channel approaches which showed higher variance.

Given the class imbalance in the dataset ($\approx$80\% normal vs.\ 20\% abnormal patients), sensitivity and specificity were reported separately throughout this section, rather than relying solely on overall accuracy, precisely to avoid masking potential weaknesses in minority-class detection. The multi-input CNN with segment selection achieved both high sensitivity (0.955) and high specificity (0.976), indicating that its performance gain is not achieved at the expense of the abnormal (minority) class. A detailed, patient-level qualitative analysis of the specific murmur types and pathologies associated with misclassified cases was beyond the scope of the present study, but represents a valuable direction for future work, particularly for understanding whether specific auscultation-spot combinations are more informative for specific pathology types.

\subsection{Limitations}

Several limitations of this study should be acknowledged. First, the requirement of a complete set of recordings from all four
auscultation spots restricted the analysis to 735 of the 1,568 patients available in the CirCor DigiScope dataset; while this
restriction follows directly from the clinical procedure the model aims to replicate, it also reduces the effective sample size and means the reported results have not been validated against the excluded subpopulation. Second, the multi-input CNN architecture was compared against baseline classifiers of considerably smaller capacity; while the consistent pattern of results across all tested configurations supports the value of synchronous multi-channel fusion, a portion of the observed gain may also be attributable to the increased representational capacity of the proposed architecture, a factor not fully isolated in the present experimental design. Third, all experiments were conducted on a single dataset (CirCor DigiScope); as this remains, to our knowledge, the only publicly available dataset offering synchronous multi-spot phonocardiogram recordings, external validation on an independent multi-channel dataset was not possible at this time, and generalization to other populations and recording conditions remains to be established. Finally, the segment selection strategy was evaluated only against random selection; comparison with simpler, computationally cheaper heuristics was left for future work.

\section{Conclusion}

This paper assessed the impact of a multi-input CNN model for synchronous analysis of heart sounds from four auscultation spots to predict patient outcomes. Leveraging the CirCor Digiscope Dataset, which provides multi-channel heart sound recordings, we segmented PCG signals into three-second intervals and extracted MFCC features. A clustering-based segment selection algorithm was proposed to identify the most representative segment per auscultation spot for model input.

The multi-input CNN achieved state-of-the-art performance with 96.5\% overall accuracy, significantly outperforming baseline
approaches. Compared to the best single-channel method (XGBoost, 82.5\%), the multi-input CNN's superiority stems from its ability to integrate information from multiple anatomical perspectives. Furthermore, it surpassed the best asynchronous multi-channel approach (SVM, 87.5\%), demonstrating that synchronous analysis can capture inter-channel interference phenomena, such as murmur wave patterns, that asynchronous methods miss. As discussed in Section~\ref{Results_section}, part of this gain may also reflect the greater representational capacity of the multi-input architecture relative to the baseline classifiers, a factor that future work should seek to disentangle from the effect of synchronous fusion itself.

The segment selection process contributed a 3.7 percentage-point performance gain over random selection (92.8\% vs.\ 96.5\%), a difference confirmed to be statistically significant by a paired Wilcoxon signed-rank test ($p = 0.003$) and a paired $t$-test ($p = 0.004$), with a moderate effect size (Cohen's $d = 0.49$), underscoring the importance of using representative, low-noise segments rather than arbitrarily chosen ones. Notably, even without segment selection, the multi-input CNN (92.8\%) substantially outperformed all alternatives, confirming the inherent advantage of synchronous multi-channel analysis. Throughout all experiments, both sensitivity and specificity were monitored to confirm that the observed gains were not achieved at the expense of the minority (abnormal) class.

In summary, this work validates three key hypotheses: (1) multi-channel approaches outperform single-channel methods, (2) synchronous analysis yields better results than asynchronous processing, and (3) segment selection positively impacts model performance, an effect supported by statistical testing rather than by cross-validation variability alone. These findings were obtained on 735 patients from a single dataset and under the experimental constraints discussed in Section~\ref{Results_section} and detailed in the Limitations, and should be interpreted within that scope.

Future work will explore multi-input architectures with heterogeneous feature types, investigate attention mechanisms to optimize segment selection and further simplify the model, and seek to disentangle the contribution of model capacity from that of synchronous multi-channel fusion, for instance through a matched-capacity single-input baseline. External validation on additional multi-spot datasets, as they become available, would further strengthen the generalizability of these findings.

\vspace{0.9em}

\textbf{Author Contributions} Marcelo Nogueira: Conceptualization, methodology, investigation, data analysis, formal analysis and writing original draft preparation; Prof. Jorge H. Oliveira, Prof. Carlos G. Ferreira, Prof. Miguel T. Coimbra and Prof. Alípio M. Jorge: Conceptualization, methodology, review and editing.

\vspace{0.9em}

\textbf{Funding} This work is financed by National Funds through the Portuguese funding agency, FCT - Fundação para a Ciência e a Tecnologia, within project LA/P/0063/2020.

\vspace{0.9em}

\textbf{Data Availability} The public CirCor DigiScope Dataset used in this study are available on their original paper and website. The code used to implement the segment selection algorithm and the multi-input CNN model described in this work is available from the corresponding author upon reasonable request.

\section*{Declarations}

\textbf{Competing interests:} The authors declare no competing interests.

\vspace{0.9em}

\textbf{Ethics approval} The research conducted for this paper adheres to ethical principles and guidelines concerning the utilization of publicly available datasets. The datasets employed in this study, The CirCor DigiScope Dataset are publicly accessible resources without individual identifiers, thus obviating the need for specific consent from individuals.

\vspace{0.9em}

\textbf{Clinical trial number} Not applicable.

\bibliographystyle{unsrtnat}
\bibliography{references}  %%% Uncomment this line and comment out the ``thebibliography'' section below to use the external .bib file (using bibtex) .





% \end{thebibliography}

\end{document}